\newcount\ShowAppendix
\documentclass[letterpaper, 10 pt, conference]{ieeeconf}
\usepackage[utf8]{inputenc}

\usepackage{url}
\usepackage{cite}
\usepackage{verbatim}
\usepackage{subcaption}
\usepackage{graphicx}
\usepackage{xcolor}
\usepackage{makecell}
\usepackage{balance}

\IEEEoverridecommandlockouts
\usepackage{amsmath}
\usepackage{amssymb}

\begin{document}

\title{\LARGE \bf Towards an Autonomous Test Driver: High-Performance Driver Modeling via Reinforcement Learning}
\author{John Subosits, Jenna Lee, Shawn Manuel, Paul Tylkin, and Avinash Balachandran
\thanks{John Subosits, Jenna Lee, Shawn Manuel, Paul Tylkin, and  Avinash Balachandran are with Toyota Research Institute, Los Altos, CA 94022, United States of America (e-mail: john.subosits@tri.global, jenna.lee@tri.global, shawn.manuel@tri.global, paul.tylkin@tri.global, avinash.balachandran@tri.global)}
}
\maketitle

\begin{abstract}
Success in racing requires a unique combination of vehicle setup, understanding of the racetrack, and human expertise. 
Since building and testing many different vehicle configurations in the real world is prohibitively expensive, high-fidelity simulation is a critical part of racecar development. 
However, testing different vehicle configurations  still requires expert human input in order to evaluate their performance on different racetracks. 
In this work, we present the first steps towards an autonomous test driver, trained using deep reinforcement learning, capable of evaluating changes in vehicle setup on racing performance while driving at the level of the best human drivers. 
In addition, the autonomous driver model can be tuned to exhibit more human-like behavioral patterns by incorporating imitation learning into the RL training process.
This extension permits the possibility of driver-specific vehicle setup optimization.
\end{abstract}

\section{Introduction}

The development of algorithms and architectures for autonomous racing is a growing research field \cite{ICRAAutonomousRacingWorkshop, betz2022autonomous}.
This growth is motivated primarily by the applicability of  approaches capable of using a vehicle's entire performance envelope to autonomous vehicle safety \cite{funke2012up, kegelman2018learning, indy_autonomous, formula_driverless, roborace, f1tenth}.  
Since autonomous motorsports has borrowed much from traditional motorsports in terms of test platforms and modeling approaches, it is natural to ask whether autonomous motorsports research has matured sufficiently to offer new insights to traditional motorsports practice.
One possible avenue for such a transfer is simulation-based development.
Control algorithms that can drive as well as and in the same style as professional drivers could stand in for them during vehicle development and tuning processes, greatly increasing the rate at which the performance of possible vehicle setups can be explored, yielding better vehicles at a lower cost.

However, such algorithms need to be able to drive at the vehicle's performance limits which are difficult to characterize \emph{a priori}.
This is particularly true when these limits vary in response to changes in vehicle design and setup.
To be able to tease out subtle differences in performance between two vehicle configurations, vehicle control algorithms must demonstrate the same ability to adjust their driving to changes in their vehicle's handling that professional human drivers do.
In the context of vehicle setup optimization for racing, this means, specifically, changes in ``grip'' and ``balance.''
Colloquially, grip refers to the overall performance potential of the vehicle -- its ability to generate accelerations for accelerating, braking, and cornering -- while balance refers to the relative ability of the front and rear axles to generate cornering forces.
A vehicle limited by the grip available from its front axle is said to be understeering, while one limited by its rear axle is said to be oversteering.
The balance can change throughout a corner and between corners, making the best vehicle setup a compromise \cite{smith1978tune}.
A well-tuned vehicle offers superior grip to support fast laps while maintaining an overall balance the driver is happy with.
Beyond this, professional drivers have distinctive driving styles when it comes to vehicle setups and choice of racing lines \cite{kegelman2018learning}, so from the practical perspective of a race engineer, it is sometimes more important to construct a driver model that imitates a specific driver than an idealized superhuman performer. 
Essentially, the relevant development goal is not the fastest car, but the fastest car \emph{for the driver who will be driving it}.

This letter presents the first empirical study of a model-free reinforcement learning agent designed to evaluate various vehicle setups.
By training the agent in a multi-task setting and conditioning the policy on the relevant vehicle parameters, we demonstrate that the agent is able to exceed the lap time performance of a professional human driver while correctly predicting the performance trend across the various vehicle setups. 
Furthermore, training one policy for all settings is less computationally expensive than individually training a policy for each.
While this approach requires no human demonstrations,  we also show that it can be extended with additional imitation learning objectives.
Incorporating human demonstrations into training data and an additional imitation loss into the policy update significantly reduces the difference between the learned policy and that implied by the human demonstrations.
Furthermore, an adjustment to the structure of the reward function allows the agent to drive more human-like trajectories as measured by differences in driven path. 
Even with the inclusion of these additional objectives, the lap time performance of the approach is able to match or exceed that of a professional racing driver in a high-fidelity motorsports simulator.


\section{Related Work}

The majority of approaches to autonomous racing and motorsports-focused driver modeling are based on traditional planning and control approaches.
For example, trajectory optimization with various levels of vehicle fidelity has been used to generate optimal trajectories that minimize lap time for a given course \cite{perantoni2014optimal,subosits2021impacts}.
These trajectories can then be tracked by state feedback controllers \cite{kapania2015design} or model predictive control.
Model predictive control allows a degree of online replanning which can partially mitigate predicted and observed tracking failures \cite{funke2016collision, liniger2015optimization, laurense2021long}.
This approach can achieve good performance \cite{kegelman2018learning, srinivasan2021holistic}, but generally falls short of the best human drivers due to modeling inaccuracies.

Some previous approaches attempt to address this shortcoming by incorporating machine learning elements into traditional control frameworks.
L{\"o}ckel \emph{et al.} used learned trajectory representations and control policies to imitate the behavior of racing drivers for motorsports simulation \cite{lockel2020probabilistic}, but their approach does not consider lap time minimization directly, though it can be combined with existing methods for speed profile adaptation from motorsports practice \cite{lockel2022adaptive}.
Spielberg \emph{et al.} used learned vehicle dynamics models to improve trajectory tracking with unknown surface friction using both feedforward \cite{spielberg2019neural} and nonlinear model predictive control \cite{spielberg2021neural} approaches.
Williams \emph{et al.} combined retraining of learned models with model predictive path integral control to create a model-based reinforcement learning approach \cite{williams2017information}.
Kabzan \emph{et al.} \cite{kabzan2019learning} further expanded on the use of learned models in MPC by incorporating uncertainty prediction to help the controller avoid constraint violations.
Moving beyond direct dynamics prediction, Rosolia and Borrelli \cite{rosolia2019learning} additionally learned value function and safe terminal sets for use with an MPC approach based on quadratic programming.
While learning can improve the performance of these approaches compared to purely physics-based models, they are not competitive with the best human drivers.

In contrast, recent simulation-based demonstrations have shown that it is possible for fully learned policies to match the performance of the best human drivers. 
Remonda \emph{et al.} explored learning from human demonstrations, without directly considering the possibility of imitating actual drivers \cite{remonda2024simulation}.
Wurman \emph{et al.} used state-of-the-art model-free reinforcement learning approaches to outperform champion drivers in both time trial and racing scenarios within the \emph{Gran Turismo Sport} video game.
However, their approach requires training a policy for each combination of car and track from scratch \cite{ wurman2022outracing}.  
An earlier study of a similar method showed that lap time can actually increase slightly when tire grip is increased due to the hyper-specialization of the agent to its environment \cite{fuchs2021super}, exactly the wrong behavior for an autonomous test driver.
In contrast, human drivers are able to adjust their driving to altered operating conditions by adjusting their implicit model of the vehicle's dynamics \cite{macadam2003understanding}.


\section{Problem Formulation and Approach}
In this work, the task of driving around the racetrack as quickly as possible is modeled as a standard Markov decision process (MDP).  
We consider each MDP as drawn from a set $M$ and specified as a tuple $\langle$$S, A, R, T, \gamma, \rho_0$$\rangle$ where $S$ is the state space, $A$ is the action space, $R(s, a)$ is the reward function, $T(s' | a, s)$ is the state transition model, $\gamma$ is the discount factor, and $\rho_0(s)$ is the initial state distribution.
The various MDPs in $M$ differ in the state transition model $T$ (and trivially in $\rho_0(s)$).
While the structure of the underlying vehicle and track models is fixed, critically for this work, the parameters that govern their behavior vary across $M$.
All other components are consistent across the various MDPs and can be engineered to produce the desired agent behavior.

\subsection{State and Action Spaces}
\begin{figure*}%
\centering
\begin{subfigure}{.32\textwidth}
\includegraphics[trim={1.5cm 1cm 0.5cm 0.5cm},clip,width=\columnwidth]{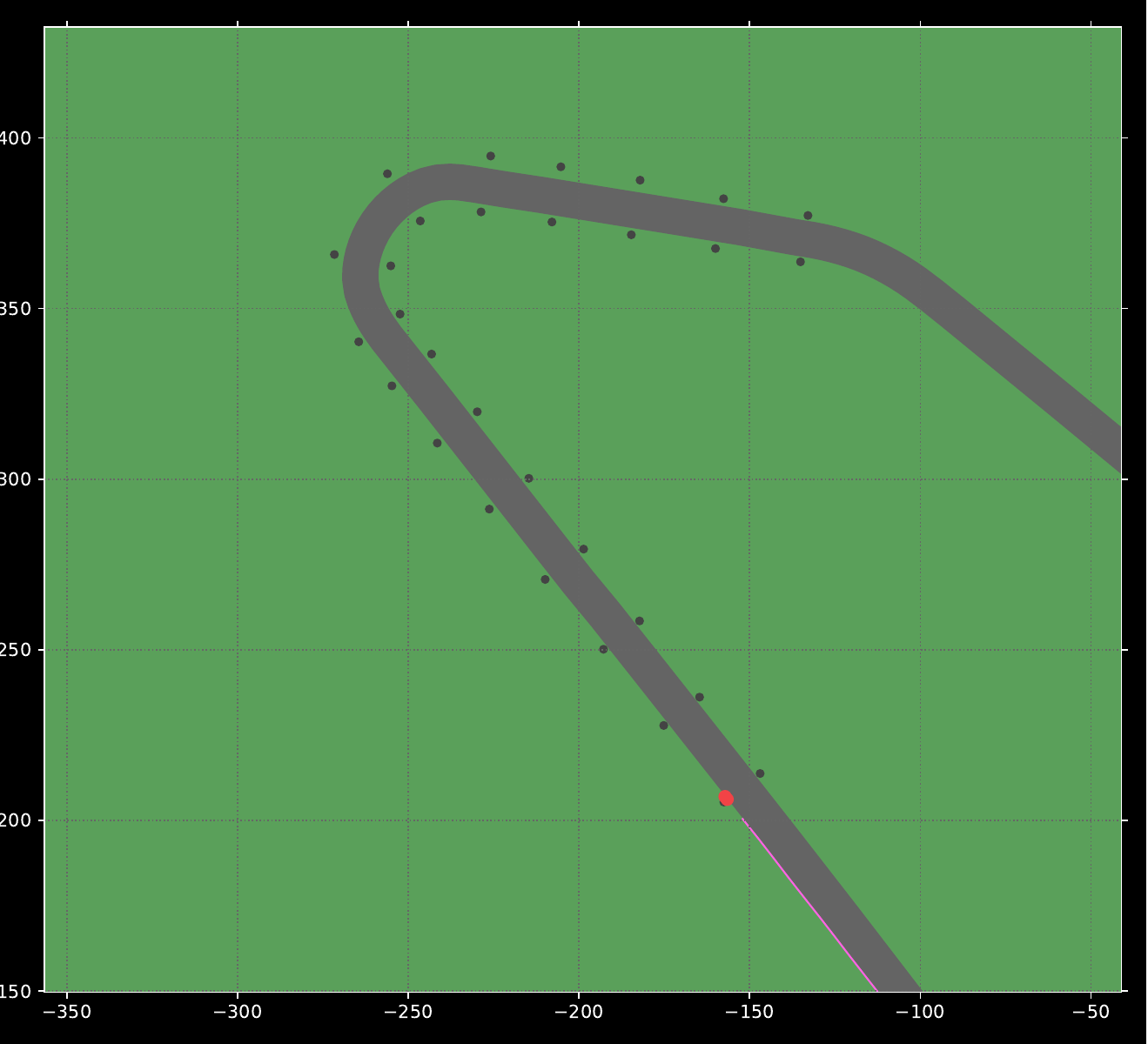}    \caption{Visualization of the agent's view of the track ahead, plotted as black points. The vehicle's prior path and current position are shown in red.}
    \label{fig:track_view}
\end{subfigure}
\hfill
\begin{subfigure}{.6\textwidth}
    \centering
    \includegraphics[width=\columnwidth]{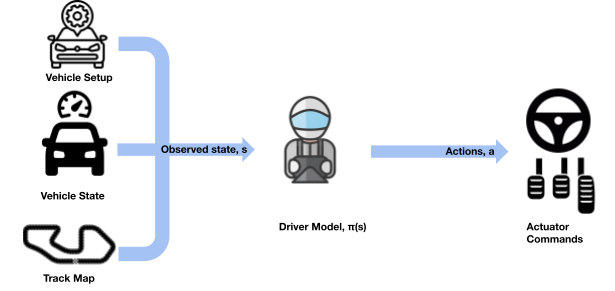}
    \caption{Schematic of the observation shared with the agent, emphasizing the importance of including the vehicle setup information.}
    \label{fig:obs_space}
\end{subfigure}%

\caption{Illustration of the state space and its components.}
\label{fig:observations}
\end{figure*}

The state and action spaces were chosen to allow the agent to capture the main characteristics of human racing driver behavior.
The observed state vector includes the vehicle's longitudinal, lateral, and yaw velocity; current steering, throttle, and brake inputs; and longitudinal, lateral, normal, and yaw accelerations.  
We also include a measure of the vehicle's lateral position on the track $e_{norm}$, with 1 and -1 corresponding to the left and right edges respectively, and the angle between the vehicle's direction of travel and the tangent to the track center line.
Information about the track ahead is included as Cartesian coordinates, in the vehicle frame, of 15 points of the left and right track edges, corresponding to roughly a 5 second horizon as shown in Fig. \ref{fig:track_view}.
Finally, a set of vehicle parameters that span the relevant vehicle setup space is included.
These three categories of observations are illustrated in Fig. \ref{fig:obs_space}.
For numerical stability, the state vector is normalized by dividing each element by its expected maximum absolute value before passing it to the agent.

A two-dimensional action space is chosen to capture steering and throttle/brake inputs.
This space is bounded to [-1, 1] by means of hyperbolic tangent functions.
Maximum steering input is mapped to roughly 30\% more than observed in human demonstrations at any circuit.
A longitudinal input of -1 is mapped to the maximum brake pedal pressure observed in human demonstrations, while a value of 0.5 maps to 100\% throttle. 
This asymmetry captures the fact that the throttle pedal is pushed against a physical stop at full throttle and makes it possible for the agent to consistently use 100\% throttle like humans do.
In this system, the agent cannot use both the brake and throttle simultaneously.
However, human drivers typically only do this very briefly. Gear shifting is handled automatically based on vehicle speed.

\subsection{Reward Function Design}
The primary goal of the reward function is to incentivize the agent to drive the fastest possible lap while observing the track limits.
The primary reward $R_s$ is meters of progress along the track centerline per time step (only given when the vehicle is on the racing surface).
A penalty of the form 
\begin{equation}\label{edge_eqn}
    R_{edge} =  q_1 e_{norm} v ^ 2, 
\end{equation}
where $v$ is the vehicle's velocity and $q_1$ is a weighting parameter used to further penalize leaving the track when $|e_{norm}| > 1$.
For consistency between the agent and human drivers, the road edges were defined as all 4 tires inside the white lines at the track edges, plus any additional space used in the available human demonstrations.

In addition to minimizing lap time, a racing driver must limit tire wear, allowing the average lap time to be minimized between tire changes.
This requirement was captured by placing a linear penalty on tire slip angles that exceeded a threshold, approximating work done by sliding friction as a proxy for tire wear,

\begin{equation}
\label{slip_eqn}
    R_{slip} =  q_2 v (\alpha_{f,\,excess} + \alpha_{r,\,excess}),
\end{equation}
where $\alpha_{i,\,excess},\ i \in \{f, r\}$ is the amount by which the slip angle exceeds the threshold for that axle.
The thresholds for the front and rear tires were set to be slightly larger than the slip angles needed to achieve maximum cornering force from that axle.

The final cost components linearly penalize steering rates and changes in longitudinal input beyond the maximum observed in human demonstrations.
The speed with which the throttle was released or the brake pedal applied was not penalized to allow as rapid of a transition from gas to brake at the end of a straightway as possible.
The complete reward function is then given by
\begin{equation}
    R = R_s + R_{edge} + R_{slip}  + R_{steer} + R_{pedals}.
\end{equation}

\subsection{Policy Learning}
The model-free reinforcement learning implementation used to learn a policy $\pi(s)$ is based on the Distributional Soft Actor Critic (DSAC) framework \cite{ma2020dsac}. DSAC is an extension of the Soft Actor Critic (SAC) \cite{haarnoja2018soft} algorithm, where the value function models the entire value distribution instead of just the mean value. 
The value function for DSAC outputs values for a fixed number of quantiles, each representing an equal probability segment in a cumulative probability density function. 

Training in a multi-task environment necessitated a population of vehicle setups with varying parameters. In each training episode, the next vehicle setup was selected from a circular queue. The vehicle was initialized in a random position on the track.
While not critical to the outcome, the initial direction of travel was always aligned with the track center line, and the initial speed was set to half of what it would be on a full speed lap. Episode length was set as approximately 25\% more than the expected lap time, but training episodes were terminated early, with an additional penalty, if the vehicle drove too far off the track or came to a stop.

\section{Experimental Results}
For purposes of comparison to a human's ability to adapt to changes in a vehicle's limit and cornering balance, the agent was trained on each of the four vehicle setups simultaneously in multiple parallel simulators. 
The simulation environment used in these experiments is a proprietary, high-fidelity multi-body simulation of the Lexus RC F GT3 racing car including the driver aids (i.e., Anti-lock Braking System (ABS) and Traction Control (TC)) present on the real car.
The track surface is modeled as a three-dimensional mesh.  
This simulation model is typically used  as part of preparation for competition, both to familiarize professional drivers with the car and the track and to assess the predicted sensitivity of vehicle performance to setup changes.

The simulation model is implemented in Dymola and was supplied to the authors as a Functional Mock-up Unit which was then incorporated into a custom OpenAI Gym environment.
The underlying simulation model must be simulated at 2000 Hz to accurately capture the high-frequency vehicle dynamics.
Despite this requirement, the agent interacts with the environment at only 10 Hz, similar to the average frequency of human inputs.
Experiments were performed on a single workstation with an NVIDIA RTX 3080ti GPU and Intel Xeon processor or an AWS EC2 g4dn.8xlarge instance with a training run taking two and six days respectively.
All simulated laps analyzed here are from Fuji Speedway with similar, unpresented, results achieved for Sebring International Raceway.

Relative to a baseline vehicle setup correlated to real world data, the other three setups were defined as an understeer condition with reduced front tire grip relative to the baseline, an oversteer condition with reduced rear tire grip relative to the baseline, and a faster condition with increased front and rear tire grip relative to the baseline.  
For testing, all four cases were given to a professional human driver to evaluate in a driver-in-the-loop simulator with motion base with which he was extremely familiar.
Compared to other possible setup adjustments, direct parameter adjustment allows the results of testing to be easily evaluated for the correct performance trend.
An additional consideration was ensuring that the human driver could take a sufficient number of laps with each setup in an acceptable amount of simulator time.

The agent is able to outperform the human demonstrations in all cases.
For evaluation during training, a snapshot of the agent's policy was tested against all four cases.
Furthermore, the vehicle was started in a single fixed position in all cases (distinct from the random initial point within a  distribution used during training) and the policy, stochastic during training, was evaluated in a deterministic manner. As a robustness check, we also evaluated the policies stochastically, which showed slower laptimes, consistent with prior work on the performance benefits of deterministically choosing the best action at each timestep. 
A summary of the human and agent performance is shown in Table \ref{table:1}.
Both the professional driver and the agent demonstrate the expected lap time trend: the baseline setup is faster than both the understeer and oversteer setups, but not as fast as the faster tires setup. As discussed further in the Appendix, similarly-strong performance was also observed by evaluating the trained agent over 100 randomly-sampled interpolated (novel to the agent) setups in the convex hull of these test conditions. 

\begin{table}[h]
\centering
\begin{tabular}{|c c  c |} 
 \hline
 \rule{0pt}{2ex}    
 Test condition & Best human lap  (s)  & Best agent lap (s)   \\ [0.5ex] 
 \hline\hline
 \rule{0pt}{2ex}    
 Baseline & 98.38&  96.63$\pm$0.02   \\ 
 \hline
 \rule{0pt}{2ex}    
 Understeer & 99.06  & 97.46$\pm$0.05   \\
 \hline
 \rule{0pt}{2ex}    
 Oversteer & 99.13  & 96.90$\pm$0.02   \\
 \hline
 \rule{0pt}{2ex}    
 Faster tires & 96.82  &  95.76$\pm$0.10  \\ 
 \hline
\end{tabular}
\caption{Comparison of human and agent lap times. Agent results are reported across four random training seeds.}\label{table:1}
\end{table}

\subsection{Analysis of Agent's Driving Across Setups}
\begin{figure}
    \centering
    \includegraphics[width=\columnwidth]{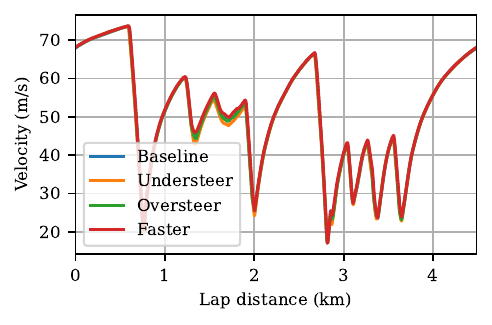}
    \caption{Comparison of agent's speed in each of the four tests showing primarily subtle differences but a qualitatively correct trend.}\label{fig:speed}
\end{figure}

A comparison of the laps driven by the agent shows subtle but distinct differences in the approach taken with each setup.
While the path taken around the track changes little, there are small but noticeable differences in the vehicle speed that account for the difference in lap time.
The differences in speed are visualized in Fig \ref{fig:speed}.
While the differences in speed are subtle, the understeer case generally has the lowest velocity and the faster case the highest, with the baseline and oversteer cases falling between them. 
In particular, the oversteer case is often faster than the baseline case on corner entry, while the reverse is true on corner exit.
Both these observations are consistent with the fact that the baseline case is generally biased towards limit understeer, that is being limited by front tire grip.
For a rear wheel drive race car such as the vehicle simulated here, understeer on corner entry and oversteer on corner exit are typical \cite{smith1978tune}.
Together these observations suggest that the back-and-forth behavior between the oversteer and baseline cases is to be expected and that the oversteer case may theoretically be faster than the understeer case even if it is difficult for a driver to consistently achieve this with the unstable car.

\begin{figure}
    \centering
    \includegraphics[width=\columnwidth]{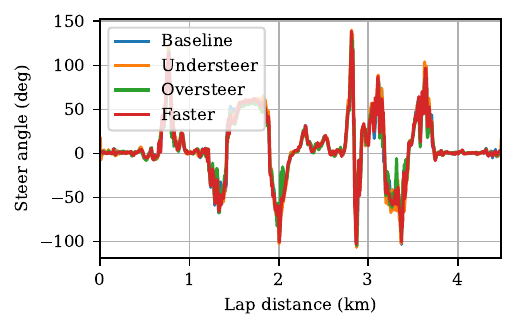}
    \caption{Comparison of agent's steering input in each of the four tests showing adjustments for oversteer and understeer.}
    \label{fig:steer}
\end{figure}

\begin{figure}
    \centering
    \includegraphics[width=\columnwidth]{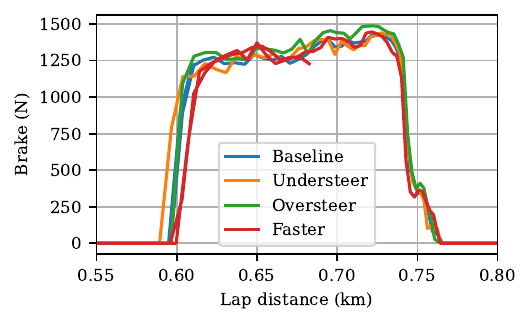}
    \caption{Comparison of agent's brake point for Turn 1 with changing friction conditions.}
    \label{fig:brake}
\end{figure}

Differences also appear in the control inputs.
As shown in the steering angle comparison in Fig. \ref{fig:steer}, more frequent and pronounced steering corrections are required in the oversteer case where the agent must quickly react to maintain directional stability.  
A similar trend was seen in rapid reductions in throttle during corner exit, suggesting that the agent is using both inputs together.
Additionally, in all but one corner, the expected trend in brake points is seen.
That is, when driving the understeering car, the agent brakes soonest, as measured by position along the track.
This is visualised for Turn 1 of the track in Fig. \ref{fig:brake}.
The difference in brake point is slight, at most 3 meters between the faster case and the baseline and another 3 meters between the baseline and the understeering case, but this shift and increased braking duration is enough to ensure that the correct trend in minimum corner speed is achieved despite the variation in braking grip across the four test cases.

\subsection{Comparison to Human Driving}
\begin{figure}
    \centering
    \includegraphics[width=\columnwidth]{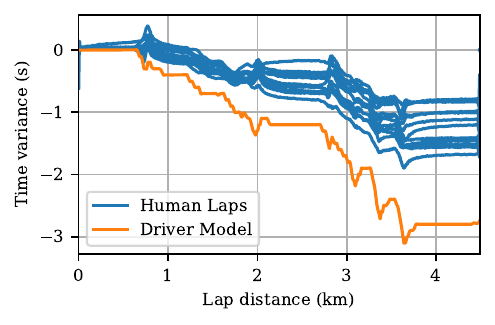}
    \caption{Gain of time relative to the human driver for the fastest setup where the difference is smallest.}
    \label{fig:delta_t}
\end{figure}

While the agent drives at an extremely high level of performance as measured by the lap times achieved, it does so by driving in a slightly different style than the human driver.  
Compared to a collection of demonstrations from the human driver, the agent drives similarly over the first half of the lap.  
As shown in Fig. \ref{fig:delta_t}, the elapsed time for both the human and the agent generally trend monotonically downward relative to a single fixed reference lap.  
A comparison of vehicle speed shows that the agent's speed mostly lies within the distribution of the human's over this part of the lap.

\begin{figure}
    \centering
    \includegraphics[width=\columnwidth]{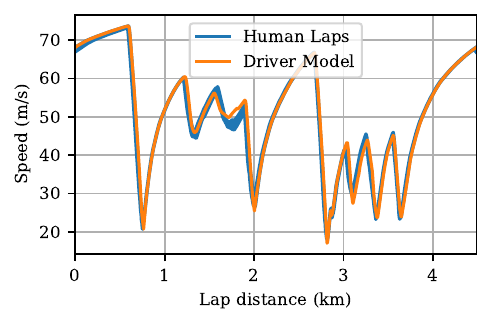}
    \caption{Comparison of agent's  speed against that of the human driver.}
    \label{fig:speedvshuman}
\end{figure}

\begin{figure}
    \centering
    \includegraphics[width=\columnwidth]{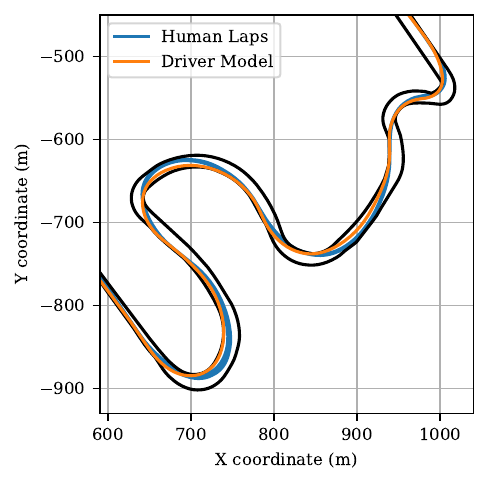}
    \caption{Comparison of racing line driven by the agent with those of the human driver. }
    \label{fig:racing_line_comp}
\end{figure}

However, significant differences appear in the latter part of the lap.
While the elapsed time advantage of the agent continues to increase faster than the human driver's, a qualitative difference is present.
Specifically, the agent is making up a substantial amount of time in first part of the corner, but giving away a portion of it during acceleration towards the next corner.
The speed profiles in Fig. \ref{fig:speedvshuman} also show a corresponding difference in behavior with the agent braking later and decelerating more aggressively, but ultimately slowing to a lower minimum speed.
Part of the reason for the significant difference between the human driver and the agent can be seen in the differences between racing lines.  
As can be seen in Fig \ref{fig:racing_line_comp}, the agent drives nearer the inside edge of the track, reducing the distance that needs to be traveled. 
An obvious downside is that the minimum speed in each corner must be less, but overall this is an advantageous trade off for the agent.

\section{Imitation of Human Demonstrations}
In previous sections, our discussion was focused on training an agent to drive as quickly as possible while paying attention to other challenges facing a racing driver. 
In this section, we present the results of methods for targeting two notions of similarity between the policy learned via RL and expert demonstrations, assuming such demonstrations are available.
An obvious measure of similarity is the likelihood that the stochastic policy learned via RL produces the same actions along the trajectory recorded from an expert demonstration.
We will refer to this as  \textbf{open-loop policy similarity}.
This supervised learning objective is the basis for behavioral cloning (BC), a straightforward approach to imitation learning.
However, unless there is extensive coverage of the state space by the demonstrations (unlikely given the precision of the best human drivers), such policy similarity is typically not sufficient to provide similarity of driven trajectories.

As an alternative, we encourage \textbf{closed-loop similarity} in the learned policy by adopting an augmented reward function in the RL problem. 
The augmented reward function is designed to guide the agent to drive a similar path around the track to that taken in the human demonstrations.
These two adjustments partially mitigate the potential sub-optimality and potential non-rationality of human drivers.

\subsection{Encouraging Open Loop Similarity}

To encourage open loop similarity in learned policy, we make two modifications commonly found in approaches to Learning from Demonstrations (LfD) \cite{ravichandar2020recent}: (1) guiding policy learning by directly adding human demonstrations to the replay buffer, and (2) ensuring learned policy to be similar to demonstrations throughout training by employing a supervised learning objective in the  policy loss.

Our approach most closely follows \cite{nair2018overcoming}, although a wide variations of the two modifications can be found across various problem domains \cite{nair2020awac, vecerik2017leveraging, galashov2022data, schmitt2018kickstarting, gupta2016learning, zhou2019watch, gao2018reinforcement}.
In contrast to the standard LfD setting, where the goal is typically improved training efficiency or higher performance, we examine the ability of such techniques to guide the agent's behavior to be more like that of a human driver.

To directly inject human demonstrations into training, we introduce an independent replay buffer and load transitions from human demonstrations. 
Demonstrations in this buffer are preserved throughout the entire life cycle of training.
In each training batch, we sample a fixed number of transitions from the auxiliary replay buffer and combine with the training data sampled from the main replay buffer, where online transitions are stored.

This modification resulted in only a marginal increase in similarity in the agent's actions and trajectory as we included more demonstration samples in the training data.
Also, consistent with the results in \cite{wang2016sample}, excessively increasing the ratio of demonstrations in the training minibatch quickly leads to unstable training. 

Second, we employed an imitation loss term in the policy loss to directly encourage the agent to take actions similar to those in the humans demonstrations. 

Formally, given that a reinforcement learning agent finds such a policy $\pi$ that maximizes the expected discounted future reward, $J$:
\begin{align*}
J(\pi) = \mathbb{E}_{(\mathbf{s}_t, \mathbf{a}_t) \sim \rho_\pi} \left[ \sum_t \gamma^t r (a_t | s_t) \right], 
\end{align*}
where $\rho_\pi(\mathbf{s}_t, \mathbf{a}_t)$ indicates state-action tuples of the trajectory when following a policy $\pi(\mathbf{a}_t | \mathbf{s}_t)$.  
As introduced in \cite{nair2018overcoming, schmitt2018kickstarting, galashov2022data}, we construct a new learning objective that incorporates the likelihood of action distribution relative to human demonstrations: 
\begin{align}
\label{eqn:Jaug}
    J_{\pi, aug} = J_\pi + \lambda_{imi} J_{imi}, 
\end{align}
where $\lambda_{imi}$ is a importance weight and imitation loss $J_{imi}$  measures the distance between the agent's learned policy and human demonstrations.
Eqn. (\ref{eqn:Jaug}) can be viewed as the standard policy loss $J_\pi$ regularized by the imitation loss $J_{imi}$, the importance of which is weighted by $\lambda_{imi}$.
This setup directly reflects the multi-objective nature of our imitation learning problem, where imitating an individual and driving the vehicle at its limits may sometimes represent competing objectives. 

We calculate the imitation loss by maximizing the likelihood of the agent's action $\mathbf{a}_t$ given the same state observations shown in human demonstrations, $\mathbf{s_t}$, which are sampled from the independent replay buffer $\mathcal{D}_{imi}$. 

Formally, this can be written as 
\begin{align}\label{eqn:Jimi}
    J_{imi} = -\mathbb{E}_{\mathbf{s}_t \sim \mathcal{D}_{imi}} \left[ \log \pi (\mathbf{a}_t | \mathbf{s}_t)\right],
\end{align}
so minimizing the augmented policy loss $J_{\pi, aug}$ is equivalent to maximizing likelihood of policy similarity provided state transitions from the demonstration. 
Typically, the value of $\lambda_{imi}$ is reduced over the course of training, allowing the agent to focus on accumulating maximum reward as its performance improves \cite{schmitt2018kickstarting}.
However, since our goal is to maintain imitative capacity throughout training, the value of this parameter was kept fixed throughout.

\subsection{Encouraging Closed Loop Similarity}
While the supervised learning approach described in the previous section captures the desire for the policy to produce similar actions to the driver, it is generally not sufficient for ensuring that rollouts of the policy look similar to human demonstrations.
While approaches such as GAIL \cite{ho2016generative} aim to address this, GAIL is not suitable for the case of multi-task learning with changing dynamics, nor has it been found to perform well in the domain of racing with fixed dynamics \cite{song2021autonomous}.
Instead, an approach used to adjust to differing dynamics between the demonstration and evaluation dynamics used the field of robot learning is adapted to the problem of imitating human driving.
This approach is to simply apply a cost penalizing differences in the paths taken during demonstrations and roll outs, \cite{atkeson1997robot, gupta2016learning}.
This approach parallels existing motorsports simulation practice which typically uses a driver model that is designed to find the fastest possible speed at which it can traverse a given, fixed path.
The similarity of the paths driven by the agent across the four vehicle setups also suggests that adjusting speed is arguably more important than adjusting the racing line when it is necessary to adjust to handling changes. 
Finally, in the human demonstrations with the varying setups, more pronounced variations in the speed over a lap were observed, supporting both our approach and traditional driver modeling  approaches.

Our implementation optionally penalizes deviations from the average path driven in the human demonstrations.
At each timestep,  an additional term of the form
\begin{equation}
    R_{imi} = -\frac{(v(u) - \mu(u))^2}{\max(\sigma(u), \sigma_{min})^2}
    \label{eq:path_track}
\end{equation}
 was added to the value of the computed reward, where $\mu(u)$ and $\sigma(u)$ are the mean and standard deviation of the lateral position of the vehicle around the track over the set of human demonstrations. The standard deviation is clipped at a minimal allowable constant value of $\sigma_{min}$ to prevent large penalties from dividing by a small standard deviation.
Meanwhile, $u$ and $v$ denote the distance along and perpendicular to the reference path defining an arbitrary curvilinear coordinate system.
Such a weighting does not guarantee that the trajectories produced by the trained agent will follow the same positional distribution as the human demonstrations.
However, it does have the desired property of incentivizing the agent to position the car precisely where the human is very consistent while permitting more variation elsewhere.

\subsection{Results}
\begin{figure}
    \centering
    \includegraphics[width=\columnwidth]{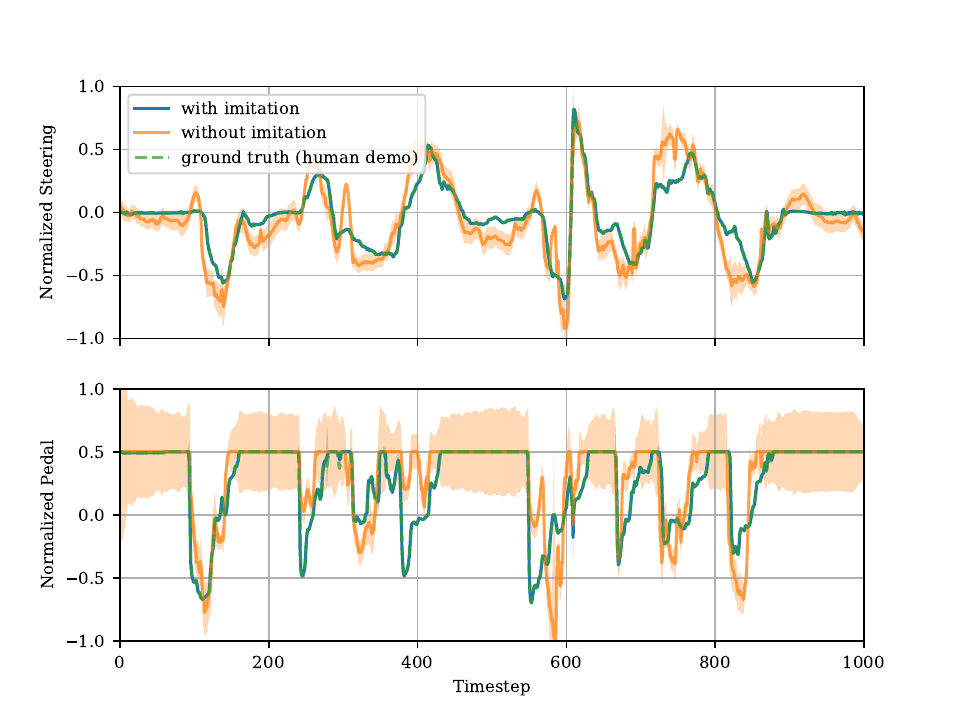}
    \caption{Open-loop comparison of policies trained with and without imitation learning, given the observations from the demonstration along the track. The dashed green line is a demonstration action taken by the human, the orange line shows actions taken by the policy trained without imitation learning, and the blue line is the policy trained with imitation learning. From these states, the agent trained with imitation learning would take essentially identical actions as the human with high certainty.}
    \label{fig:open_loop_similarity}
\end{figure}

\begin{figure}
    \centering
    \includegraphics[width=\columnwidth]{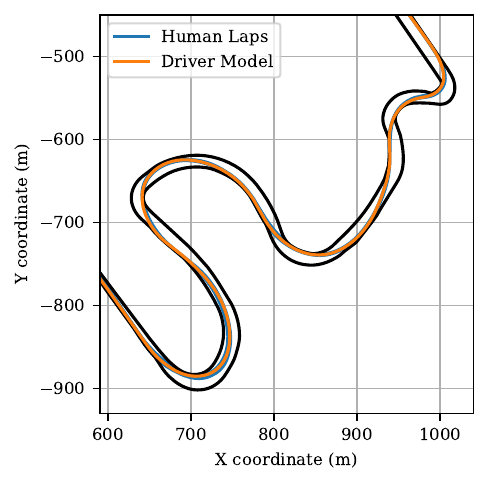}
    \caption{Comparison of agent's fastest lap with the twelve human examples available for the fastest setup of the four showing qualitatively how the difference in the driven path has been greatly reduced compared to Fig. \ref{fig:racing_line_comp}.}
    \label{fig:racing_line_imitation}
\end{figure}

The results of experiments with these approaches to driver imitation are summarized in Tables \ref{table:drving_style} and \ref{table:lap_time_comp}.
Sharing human demonstrations with the agent as additional experience was found to have little impact on any metric.
As expected, imitation learning loss has little impact on path tracking similarity, but is extremely effective in ensuring that the agent takes similar actions to the human driver from a given state as shown in Fig. \ref{fig:open_loop_similarity}.
In contrast, the augmented reward function is extremely effective in reducing differences in the driven path and has a modest impact on the likelihood that the policy takes similar actions.
A comparison of the racing lines is shown in Fig. \ref{fig:racing_line_imitation} for the case where both the augmented reward function and the supervised learning policy loss were used.
Interestingly, the augmented reward function, despite being always being less than or equal to the baseline reward function, led to slightly  more accumulated reward (and lower average lap time).
This suggests that the additional term  could serve as an effective means of reward shaping and that overall the human driver drives a slightly better racing line than the agent arrives at by itself.
Ultimately our results indicate that both policy and trajectory similarity can be achieved with little cost in terms of outright performance, allowing the agent to drive the vehicle to its limit while emulating a human driver.

\begin{table*}
\centering
\caption{Comparison of performance metrics across the variations on training, each reported across four random training seeds, with associated standard deviations.}
\begin{tabular}{|c c c c  c|} 
 \hline
 Test condition & \makecell{Policy Similarity \\ (log-likelihood)} & \makecell{Path Deviation \\ (Eq. \ref{eq:path_track})}  &  Average agent lap time (s) & Average Reward \\ [0.5ex] 
 \hline\hline 
\rule{0pt}{2ex}    
 Pure RL & -5.88 $\pm$ 3.0 & 85,804 $\pm$25,764 &  96.69 $\pm$ 0.06 & 6899.79 $\pm$ 6.85 \\  
 \hline
\rule{0pt}{2ex}    
 Shared experience and regularization & \textbf{4.95 $\pm$ 0.19} &  89,732 $\pm$ 30,527     & 96.72 $\pm 0.10 $ & 6898.76 $\pm$ 16.69 \\
 \hline
 \rule{0pt}{2ex}    
Regularization only & 4.62$\pm$ 0.24 & 86,405 $\pm$ 21,663  & 96.75 $\pm$ 0.07 & 6899.79 $\pm$ 6.85 \\
 \hline
 \rule{0pt}{2ex}    
 Augmented reward only& -1.09$\pm$ 0.38 & \textbf{507.68 $\pm$ 22.22} &  \textbf{96.42 $\pm$ 0.03} & \textbf{6903.34 $\pm$ 3.14} \\
 \hline
 \rule{0pt}{2ex}    
 Augmented reward and regularization & 4.91$\pm$ 0.25 & 2,842.3 $\pm$ 2179.2  &  96.54 $\pm$ 0.06 & 6840.86 $\pm$  44.12\\
 \hline
\end{tabular}

\label{table:drving_style}
\end{table*}

\begin{table*}
\centering
\caption{Comparison of the best lap time performance by setup across the variations on training.  }
\begin{tabular}{|c c c c  c|} 
 \hline
 Test condition & Baseline & Understeer  &  Oversteer & Faster tires \\ [0.5ex] 
 \hline\hline 
\rule{0pt}{2ex}    
 Pure RL &96.63$\pm$0.02 & 97.46$\pm$0.05 &  96.90$\pm$0.02 & 95.76$\pm$0.10 \\  
 \hline
\rule{0pt}{2ex}    
 Shared experience and regularization & 96.67$\pm$ 0.11 &  97.44$\pm$ 0.07     & 96.94$\pm$ 0.11  & 95.83$\pm$ 0.11 \\
 \hline
 \rule{0pt}{2ex}    
Regularization only & 96.70$\pm$ 0.09 & 97.47$\pm$ 0.06  & 97.00$\pm$ 0.11 & 95.86$\pm$ 0.1 \\
 \hline
 \rule{0pt}{2ex}    
 Augmented reward only& \textbf{96.33$\pm$ 0.04} & \textbf{97.19$\pm$ 0.05} &  \textbf{96.70$\pm$ 0.07} & \textbf{95.47$\pm$ 0.03} \\
 \hline
 \rule{0pt}{2ex}    
 Augmented reward and regularization & 96.47$\pm$ 0.03 & 97.22$\pm$ 0.04  &  96.74$\pm$ 0.06 & 95.71$\pm$ 0.13\\
 \hline
\end{tabular}

\label{table:lap_time_comp}
\end{table*}

\section{Conclusion}
Simulation-driven development is a key capability in motorsports and automotive engineering, but the utility of such simulations is limited by the difficulty of modeling human drivers and their ability to adjust to different vehicle setups.
This letter presents a reinforcement learning based approach to driver modeling that can be extended to include direct imitation of a human driver as a secondary objective.
Our future work will examine the degree to which inverse reinforcement learning can be used to rationalize differences between drivers' driving styles and the usefulness of meta-learning in quickly adapting the control policy to vehicle setups that are outside the distribution of those seen during training.
The experiments reported in this letter show that by training with a variety of vehicle setups and including the salient setup parameters in the state space, reinforcement learning can be used to train policies that drive lap times better than a professional human driver.
Furthermore, by adjusting the reward function and incorporating an imitation learning loss into the policy update, policies that are more similar to human demonstrations can be generated.
These polices are more similar in that they are likely to take a similar action as the human driver did in a given situation and in the fact that they generate trajectories that are more similar to those produced by the human driver.
The ability to generate policies for robustly driving vehicle setups with varying performance potentials and predicting their performance envelopes significantly aids the goal of using high-fidelity simulation to assess the sensitivity of vehicle performance to setup changes.

\section*{Acknowledgement}
We would like to thank our colleagues at Lexus Gazoo Engineering and Toyota Racing Development, especially Sunder Vaduri and Chris Kee.  
Their technical guidance and feedback has been invaluable in the preparation of this letter.

\medskip

\bibliographystyle{IEEEtran}
\bibliography{main}
\balance

\ifnum\ShowAppendix=1
\newpage
\onecolumn
\section*{Appendix}

\section{Network Architecture}
The network architecture for DSAC consists of a policy network and two quantile value networks. 
The policy network maps observations to steer and pedal actions, and the value networks map observation and actions to the values for 32 quantiles. 
DSAC uses the quantile regression loss \cite{dabney2018distributional} to fit these quantile values to the minimize the Wasserstein distance between the value function's output and the target value distribution.
The policy and two quantile networks each have a target network, whose parameter values lag behind the original networks. The pararmeters of the target networks are updated incrementally towards the original networks via a soft update parameter, $\tau$, of 0.005. Each network has three fully-connected layers with 2048 neurons per layer.

Before training, a replay buffer of max size 1,000,000 was pre-loaded with 10,000 steps of experience tuples generated using a random policy. Training alternated between a single environment sample step in 16 parallel environments and a training step where a batch size of 256 was sampled from the replay buffer to train the models. The learning rate ($\alpha$) of 0.0003 and a discount rate ($\gamma$) of 0.99 was used. Table \ref{table:0} summarizes the values of various hyperparameters used during training.

\begin{table}[h!]
\centering
\begin{tabular}{|c c |} 
 \hline
 Hyperparameter & Chosen Value\\ [0.5ex] 
 \hline\hline
 Number of quantiles & 32   \\ 
 \hline
 Hidden units & 3 layers of 2048 units each   \\
 \hline
 Target update rate ($\tau$) & 0.005   \\
 \hline
 Replay buffer max size & $10^6$   \\  
 \hline
 Replay batch size & 256   \\  
 \hline
 Learning rate ($\alpha$) & 0.0003   \\  
 \hline
 Discount rate ($\gamma$) & 0.99   \\  
 \hline
 Number of parallel simulators & 16   \\  
 \hline
\end{tabular}
\caption{Chosen hyperparameters and network architecture for each experiment}
\label{table:0}
\end{table}

\section{Reward Function Component Weights and Imitation Parameters}

Table \ref{tablereward:0} below presents the weights for each component of the reward function and the additional parameters used for the imitation learning and augmented reward function across each of the five variants of experiments run. In the table, the progress weight is the multiplier for the $R_s$ reward component, bounds exceed penalty weight corresponds to $q_1$ in (\ref{edge_eqn}). The slip angle penalty weight corresponds to $q_2$ in (\ref{slip_eqn}). Steer rate penalty weight and pedals rate penalty weight are the multiplier for the $R_{steer}$ and $R_{pedals}$ reward components respectively. 
The imitation batch \% is the proportion of the batch size which is sampled from the auxiliary buffer of human demonstrations and combined with the remaining batch proportion taken from the RL replay buffer used for training the agent. The imitation weight is the term $\lambda_{imi}$ used in (\ref{eqn:Jaug}). The imitation batch size is the size of the minibatch of samples drawn from the dataset of human demonstrations for each update step in (\ref{eqn:Jimi}). The lateral dist std floor refers to the minimum standard deviation term $\sigma_{min}$ in (\ref{eq:path_track}). The path tracking deviation penalty weight is the multiplier for the $R_{imi}$ reward component.

\begin{table}[h!]
\centering
\begin{tabular}{|c c c c c c|} 
 \hline
 Hyperparameter & Pure RL & Shared exp + reg & Reg only & Augmented reward & Augmented reward + reg\\ [0.5ex] 
 \hline\hline
 Progress weight & 1 & 1 & 1 & 1 & 1   \\  
 \hline
 Bounds exceed penalty weight ($q1$) & 1 & 1 & 1 & 1 & 1   \\  
 \hline
 Slip angle penalty weight ($q2$) & 1.5 & 1.5 & 1.5 & 1.5 & 1.5   \\  
 \hline
 Steer rate penalty weight & 16 & 16 & 16 & 16 & 16   \\  
 \hline
 Pedal rate penalty weight & 8 & 8 & 8 & 8 & 8   \\  
 \hline
 Imitation batch \% & & 0.1 & 0 & & 0 \\ 
 \hline
 Imitation weight ($\lambda_{imi}$) & & 10 & 10 & & 10  \\
 \hline
 Imitation batch size & & 256 & 256 & & 256  \\
 \hline
 Lateral dist std floor ($\sigma_{min}$) & & & & 1 & 1   \\  
 \hline
 Path tracking deviation penalty weight & & & & 0.1 & 0.1   \\  
 \hline
\end{tabular}
\caption{Reward function weights and imitation parameter values}\label{tablereward:0}
\end{table}

\section{Comparison between DSAC and baseline SAC algorithm}

To justify the use of the DSAC algorithm, we ran the Pure RL experiment using the SAC algorithm for 100 iterations and with the same network architecture without quantiles. Table \ref{table:sac_baseline} shows the maximum reward achieved as well as the lap times when evaluated on each of the four setups. Compared to DSAC, the total reward achieved by SAC was considerably lower and the resulting lap times were all higher than the human driver's best lap times for each setup.

\begin{table}[h]
\centering
\begin{tabular}{|c c c c c  c|} 
 \hline
 Test condition & Reward & Baseline & Understeer  &  Oversteer & Faster tires \\ [0.5ex] 
 \hline\hline 
\rule{0pt}{2ex}    
 DSAC & 6899.79$\pm$6.85 & 96.63$\pm$0.02 & 97.46$\pm$0.05 &  96.90$\pm$0.02 & 95.76$\pm$0.10 \\  
 \hline
\rule{0pt}{2ex}    
 SAC & 6473.12$\pm$36.15 & 100.03$\pm$ 0.3 &  100.7$\pm$ 0.71     & 100.22$\pm$ 0.28  & 98.9$\pm$ 0.78 \\
 \hline
\end{tabular}
\caption{Comparison of DSAC to SAC performance}
\label{table:sac_baseline}
\end{table}

\section{RL trained on single setup compared to all setups}

To investigate the ability of the same RL policy trained on multiple setups simultaneously compared to training on an individual setup, we ran an ablation of the Pure RL experiment with the policy learning only with the baseline vehicle setup. The difference in best achievable performance of training the RL agent in a single vehicle setup compared to training on all four setups simultaneously is shown in Fig \ref{fig:num_setups} and is asymptotically negligible, although training on all four setups simultaneously may take a few extra iterations at the start of training to reach a similar reward, this allows it to achieve an identical best lap time on the baseline vehicle as the policy trained only in that setup. 

\begin{figure}[h]
    \centering
    \includegraphics[width=0.48\columnwidth]{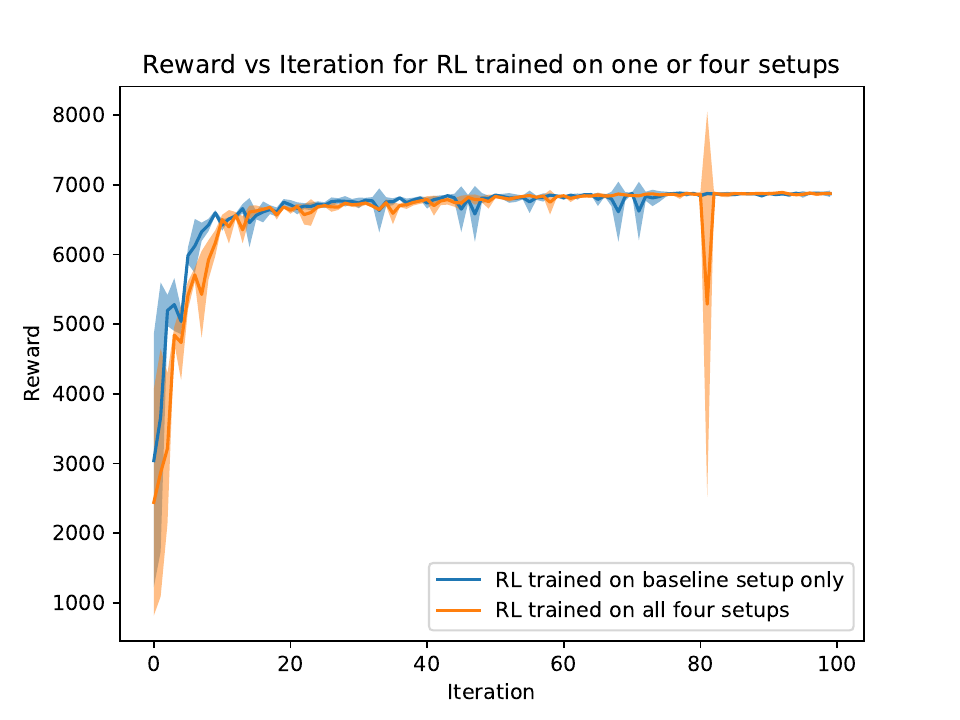}
    \includegraphics[width=0.48\columnwidth]{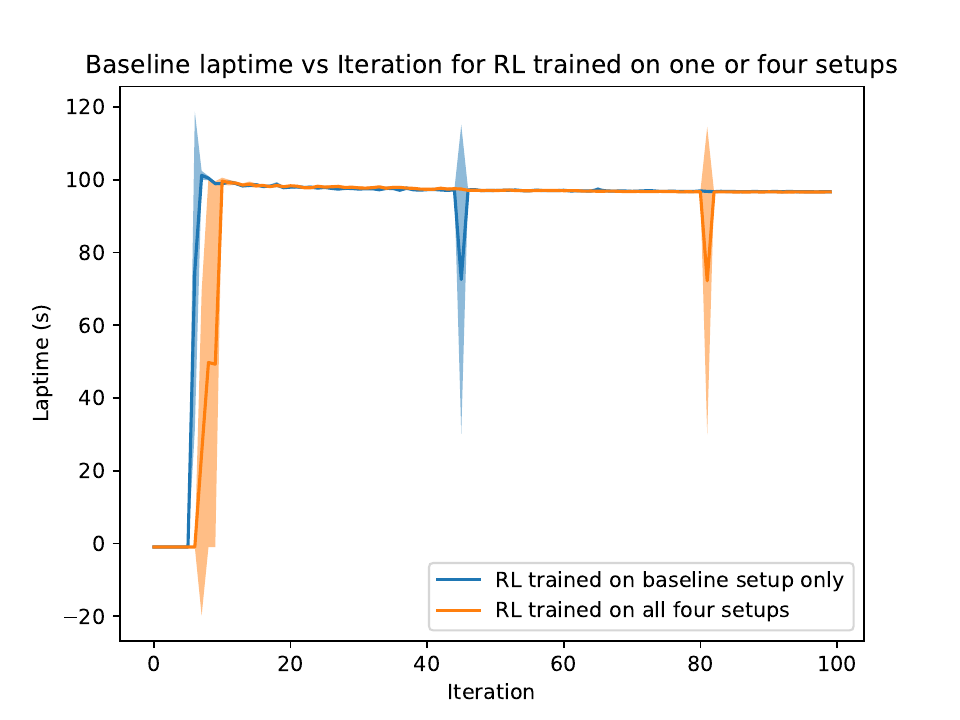}
    \caption{(left) Comparison of the average total reward at each iteration of training. (right) The lap time on the baseline setup at each iteration of training. In each graph, the policy trained on all setups simultaneously is in orange and the policy trained on the baseline setup is shown in blue. }
    \label{fig:num_setups}
\end{figure}

\section{Results with unseen setups}
To evaluate the ability of the learned policy to generalize to setups that were not seen during training, we generated 100 random vehicle setups with different tire friction scale factors.
These setups were drawn from the convex hull of the four training setups, meaning that these unseen setups are, in some sense, ``in distribution'' despite not being seen during training.
In all 100 cases, the policy was able to complete a lap despite not having any experience with the setup. 
As seen in Fig. \ref{fig:unseen_setups}, the performance trends are well described (MAE 0.020 sec) by a simple linear expression $118.65 - 25.77 \lambda_{\mu_f} - 7.00 \lambda_{\mu_r} $   where $\lambda_{\mu_f}$ and $\lambda_{\mu_r}$ are the arbitrary front and rear friction scale factors. 
This suggests that the method is able to make stable lap time predictions as the vehicle setup is varied.

\begin{figure}[h]
    \centering
    \includegraphics[width=0.48\columnwidth]{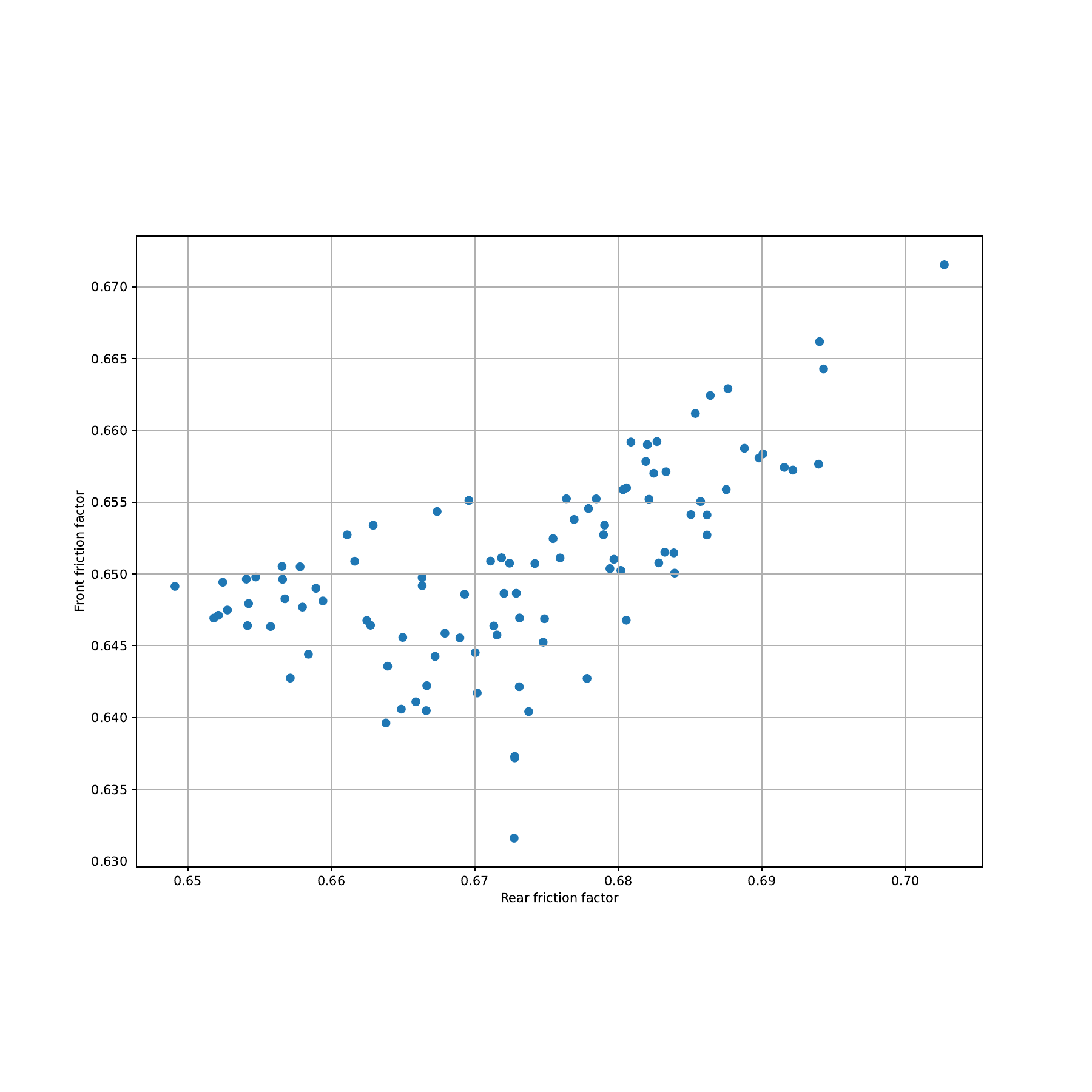}
    \includegraphics[width=0.48\columnwidth]{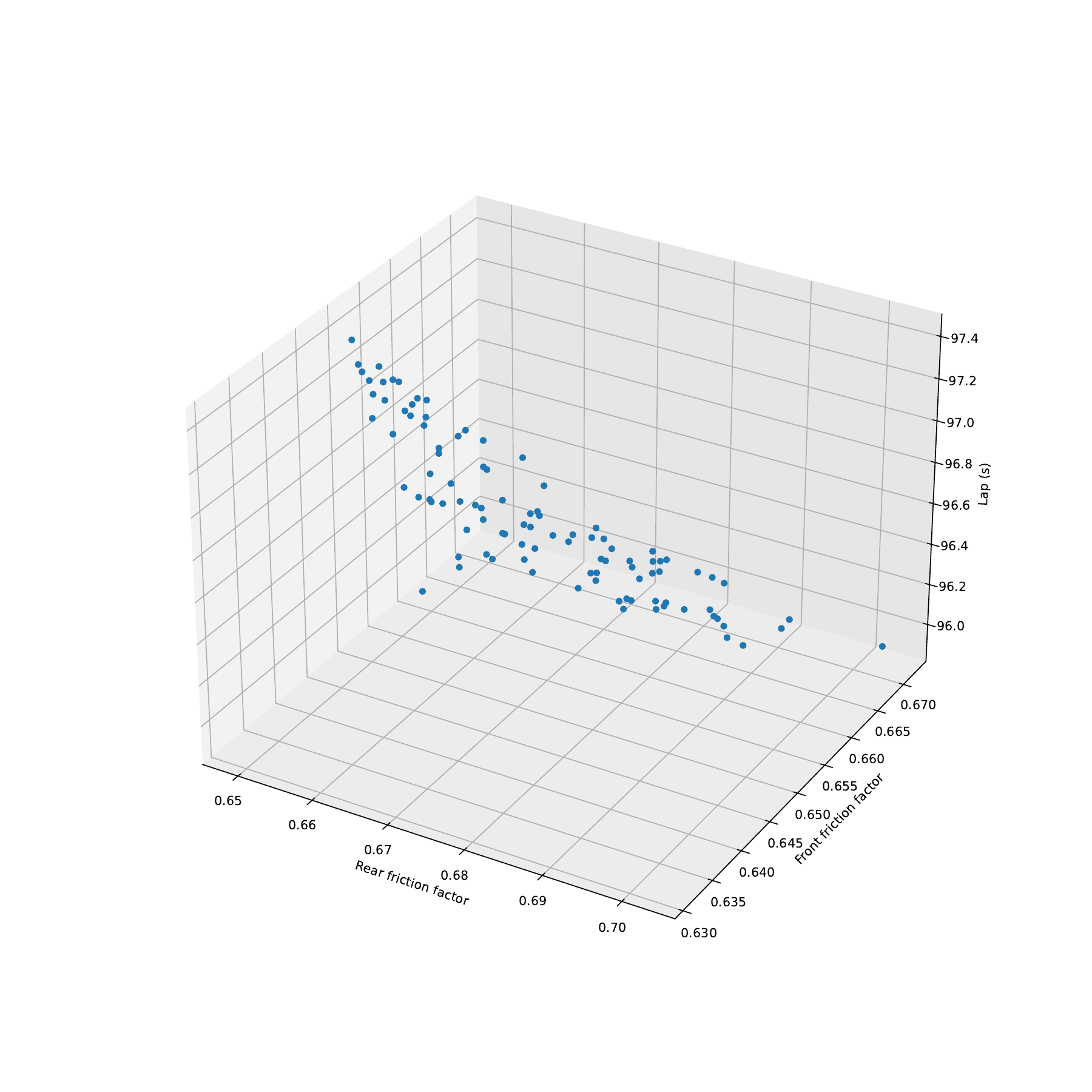}
    \caption{The lap time trend (right) is consistent across the random setups (left).}
    \label{fig:unseen_setups}
\end{figure}

\fi

\end{document}